# Feature Weighting for Improving Document Image Retrieval System Performance

Mohammadreza Keyvanpour[1], Reza Tavoli[2]

[1] Department Of Computer Engineering, Alzahra University
Tehran, Iran

[2] Department of electrical and computer, Qazvin Islamic Azad University (QIAU)
Qazvin, Iran

**Abstract**

Feature weighting is a technique used to approximate the optimal degree of influence of individual features. This paper presents a feature weighting method for Document Image Retrieval System (DIRS) based on keyword spotting. In this method, we weight the feature using coefficient of multiple correlations. Coefficient of multiple correlations can be used to describe the synthesized effects and correlation of each feature. The aim of this paper is to show that feature weighting increases the performance of DIRS. After applying the feature weighting method to DIRS the average precision is 93.23% and average recall become 98.66% respectively.

**Keywords:** *Information Retrieval, Indexing, Document Image, Feature weighting.*

## 1. Introduction

Document Image Retrieval System (DIRS) based on keyword spotting is performing the matching directly in the image data bypassing OCR and using word-images as queries. In DIRS [1], no weights don't assign to the extracted features and weights for all features is one, although some features more effect to retrieval.

Feature weighting is a feature importance ranking algorithm where weights, not only ranks, are obtained [14]. Commonly used feature weighting methods only consider the distribution of a feature in the documents and do not consider class information for the weights of the features. Several methods were reported for feature weighting to be based on such as term frequency (TF) [16], inverse document frequency (IDF) [17]. In [15], a framework for integrating multiple, heterogeneous feature spaces in the k-means clustering algorithm is presented.

In this paper, we propose a feature weighting method for increase performance of Document Image Retrieval System based on exact word matching. The proposed method weights each feature in the words according to the different role of the features during the indexing process. The aim of this paper show that feature weighting increases the performance of DIRS, In terms of average recall, average precision.

The reminder of the paper is organized as follows: section 2 surveys previous related works in document image retrieval. Section 3 describes Document Image Retrieval System. Section 4 describes the proposed system. Section 5 explains evaluation measures used in this paper. Section 6 will show the experimental results of the proposed system. Section 7 is the conclusion.

## 2. Related Works

In recent years, a number of attempts have been made by researchers to retrieval document images by word image. A detailed survey on document image retrieval up to 1997 can be found in Doermann [7]. In [10] an overview on document image retrieval system is presented. Word level image matching and retrieval has been attempted for printed documents [3, 4, 5, 6, 8, 9, 11, 12, 13].

Leydier et al [8] used DIP techniques to create a pattern dictionary of each document and then they performed word spotting by selecting the feature of the gradient angle and a matching algorithm. In previous works on Word Shape Coding, Li [5] uses an alternative technique and combination of feature descriptors for keyword spotting without the use of OCR. In [4] a novel partial matching algorithm is designed for morphological matching of word form variants in a language. In [3] using of document image processing techniques extract powerful features for the description of the word images. In [6], Shijian Lu annotates word images by using a set of topological shape features including character ascenders /descenders, character holes, and character water reservoirs. With the annotated word shape codes, document images can be retrieved by either query keywords or a query document image. In [9], Liz hangs et al proposed for designing an information retrieval system with ability of dealing with imaged document stored in digital libraries .The proposed system provide an efficient and promising tool for document image retrieval .This method is used to propose word coding techniques.

## 3. Document Image Retrieval System (DIRS)

Figure1 depicts the overall structure of the Document Image Retrieval System Base on Word Spotting [1]. This system composed of two different sections: the offline and the online operation. In the offline operation the repository of document images are examined and the results are stored in a database. This section consists of three stages. At first stage the document transports the preprocessing stage which includes a binarization with the Otsu method, a mean filter and a skeletonization operation. The word segmentation stage is following the preprocessing. Its primary goal is to detect the word blocks. This is achieved with the continuously use of vertical and horizontal projections (Recursive X-Y Cuts). In the final stage of the offline operation the features of each word are calculated and stored in the database. For each word block, a total of 7 different features in use, namely, Width to height ratio, Word area density, Center of gravity, Vertical projection, Top–bottom shape projections, Upper grid features and down grid features.

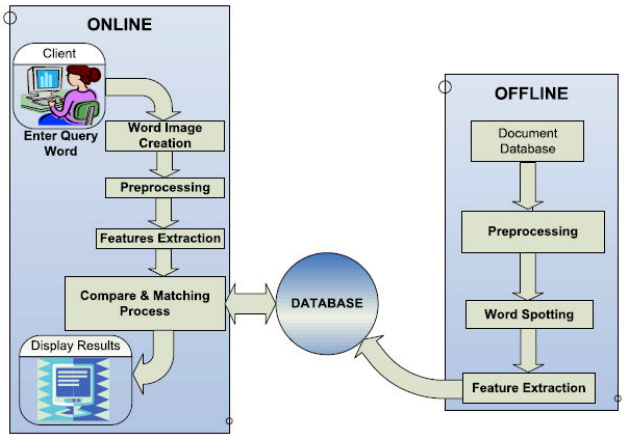

Fig 1.Overall structure of the Document Image Retrieval System Base on Word Spotting [1]

Word area density is calculated by using the following relation:

$$E = 100 \frac{BP}{(IW)(IH)} \quad (1)$$

Where (BP) is the number of black pixels in the word bounding box, (IW) is the width and (IH) is the height of the word bounding box in pixels.

In order to calculate center of gravity, the vertical and horizontal center of gravity must be determined by the following equations:

$$C_x = \frac{M_{(1,0)}}{M_{(0,0)}} \quad (2)$$

$$C_y = \frac{M_{(0,1)}}{M_{(0,0)}} \quad (3)$$

Where $C_x$ is the horizontal center and $C_y$ the vertical center of gravity and M (p, q) the geometrical moments of rank p+q:

$$M_{pq} = \sum_x \sum_y \left(\frac{x}{width}\right)^p \left(\frac{y}{height}\right)^q f(x,y) \quad (4)$$

The x and y determine the image pixels. Finally, the center of the gravity feature is defined as equal to the Euclidean distance from the upper left corner of the image:

$$COG = \sqrt[2]{c_x^2 + c_y^2} \quad (5)$$

The vertical projection is defined as the number of black pixels that they are residing in each column i. This feature consists of the first twenty (20) coefficients of the discrete cosine transform (DCT) of the smoothed and normalized vertical projection. The vertical projection of "performance" is depicted in figure 2.

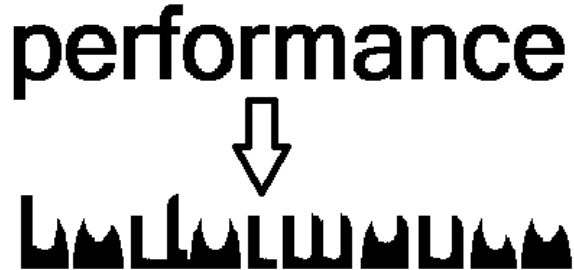

Fig 2. Vertical projection of "performance"

In order to calculate the top shape projection, the word image is scanned from top to bottom. The first time a black pixel is found all the following pixels of the same column are converted to black. The top projection of "performance" is depicted in figure 3. This feature consists of the first 25 coefficients of the discrete cosine transform (DCT) of the smoothed and normalized top projection.

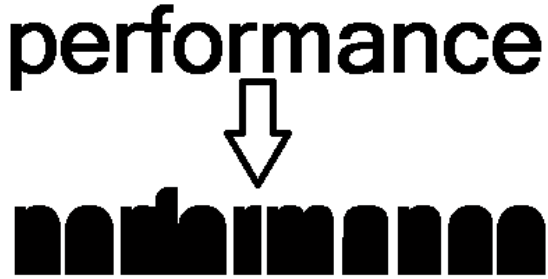

Fig 3. Top projection of "performance"

The bottom shape projection is found similarly. The word image is scanned from bottom to top and all the pixels are converted to black until a black pixel is found. This feature consists of the first 25 coefficients of the discrete cosine transform (DCT) of the smoothed and normalized bottom projection.

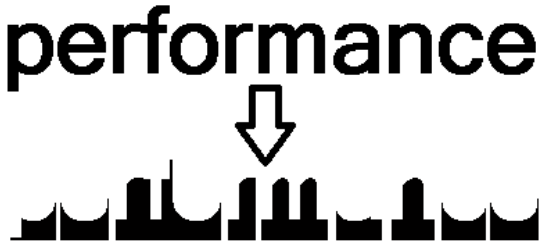

Fig 4. Bottom projection of "performance"

The upper grid features (UGF) is a ten element vector with binary values extracted from the upper part of each word image. The down grid features (DGF) is a ten element vector with binary values extracted from the lower part of each word image.

The online operation of the DIRS is the visible part from the user perspective. It consists of the web interface from which the users enter the query word and see the results. The creation of the word's image, the preprocessing and features extraction stages which are the same with that in the offline operation, and finally, the matching process of the query word's features with them in the database. The matching procedure can identify the word images of the documents that are more similar to the query word through the extracted feature vectors [1].

## 4. Overview of Feature Weighting For Document Image Retrieval System (FWDIRS)

The overall structure of proposed system is displayed in figure 5. The proposed system is designed to perform weighting features based on coefficient of multiple correlations. In fact, main difference between this system and DIRS is Feature Weighting and Similarity Measurement stage.

In the feature weighting stage, stored features in database weighted by its importance. In DIRS, no weights assign to the extracted features and weights for all feature equal one, although some features more effect to retrieval. This method weights each feature according to the different role of the features during the indexing process. Next section describes this method in details.

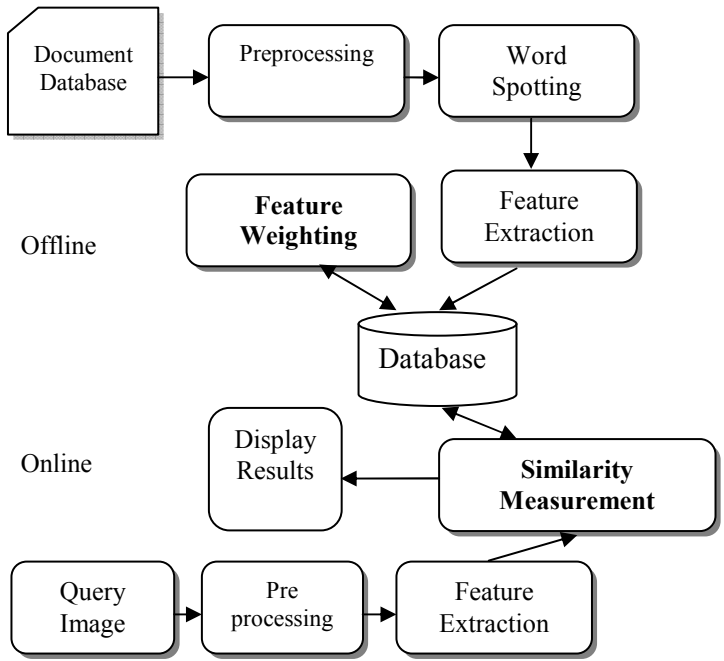

Fig 5. Proposed System

### 4.1 Feature Weighting

The basic idea of feature weighting is inspired from weighting attributes by coefficient of multiple correlations [2]. Coefficient of multiple correlations can be used to describe the synthesized effects and correlation of each attribute (Feature). The degree of multiple correlations of several attributes and one attribute are measured by coefficient of multiple correlations. Coefficient of multiple correlations can be calculated using coefficient of single correlation and coefficient of partial correlation. Let attribute y be a function and attributes $x_1, x_2, \ldots, x_n$ be variables, then the coefficient of multiple correlation between y and $x_1, x_2, \ldots, x_k$ (when there are k variables) is $R_{y,1\ 2\cdots k}$. The formula is as follows when there are k variables:

$$R_{y,12\ldots k} = \sqrt{1-(1-r_{y1}^2)(1-r_{y2.1}^2)\ldots(1-r_{yk.12\ldots(k-1)}^2)} \quad (6)$$

In which, $r_{y1}$, $r_{y2.1}$, $r_{yk.12\cdots (k-1)}$ are coefficient of single correlation and coefficient of partial correlation. If considering a selected attribute $x_i$ using the correlations between the attribute and all the other attributes - the coefficient of multiple correlations $\lambda_{x1,x2,\ldots,xk}$ which is simply represented as $\lambda_i$, it reflects the ability of those attributes other than xi to replace $x_i$. When $\lambda_i = 1$, $x_i$ can be removed or the weight of which should be decreased; when $\lambda_i$ is very small, non- $x_i$ values cannot replace it and should increase its weight. Thus, $|\lambda_i|^{-1}$ can be used to calculate the weight $w_i$:

$$W_i = \frac{|\lambda_i|^{-1}}{\sum_{j=1}^{k}|\lambda_i|^{-1}}, i=1,2,\ldots,k \quad (7)$$

$w_i$ is the absolute value of coefficient of multiple correlation such that $0 \leq w_i \leq 1$.

Assume DO= $\{do_1, do_2, do_3, \ldots, do_n\}$ is the array of data objects that stored in database. After calculate weights for each feature, weighted data objects are: $DO_i = W \cdot DO_i$, $i=1, 2\ldots n$.

$$W = \begin{bmatrix} w_1 & 0 & 0 & 0 \\ 0 & w_2 & 0 & 0 \\ 0 & 0 & w_3 & 0 \\ 0 & 0 & 0 & w_m \end{bmatrix} \quad (8)$$

4.2 Similarity Measurement

The matching procedure can identify the word images of the documents that are more similar to the query word through the extracted feature vectors. In this system weighted Minkowski distance between the features of the query image and the word included in the processing document is calculated:

$$d(do_i, q) = [\sum_{k=1}^{93} w_k |do_{ik} - q_k|], i = \{1,2,\ldots,n\} \quad (9)$$

Where $d(do_i, q)$ is the Minkowski distance of the word $i$ and query $q$, $k$ is the feature which is being compared, $q_k$ is the query descriptor, $do_{ik}$ is descriptor of word i and $w_k$ is weight of kth feature.

## 5. Evaluation Measures

Precision and recall measures are widely used for evaluation of the document image retrival system. They are defined as follows in (10) and (11):

$$precision = \frac{\#(Relevant\ Retrieved\ Records)}{totalNumberOf\ Retrieved\ Record} \quad (10)$$

$$recall = \frac{\#(Relevant\ Retrieved\ Records)}{totalNumberOf\ Relevant\ Record} \quad (11)$$

In our evaluation the precision and recall values are expressed in percentage.

## 6. Experimenting the Proposed System

In our experiments, the evaluation of the proposed system was based on 100 document images. The database of the documents has been created automatically from various digital text documents. In order to calculate the precision and recall values 30 searches were made using random words. The precision and recall values obtained are depicted in Figure 5.

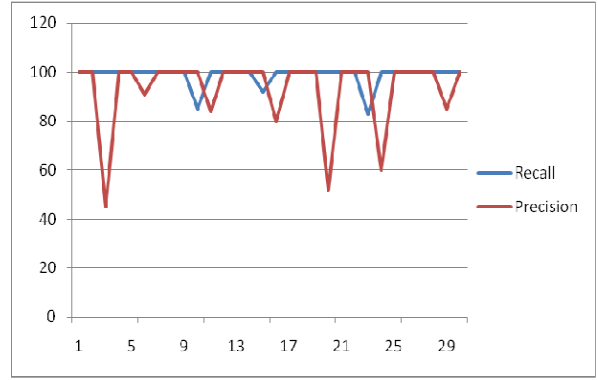

Fig 6. The variation of the precision and Recall Coefficient of the proposed FWDIRS for 30 searches. The Average Precision is 93.23% and the average Recall is 98.66%

As shown in Figure 6, by using coefficient of multiple correlations for feature weighting, performance of DIRS in term of average prcision is increased and term of average of recall is fixed.

Table 1 dpicts comparison the average precision and average recall of the approach with DIRS [1] and WDIRS [3].

Table 1. Comparison the average precision and recall between proposed system and DIRS and WDIRS

|  | Precision | Recall |
|---|---|---|
| WDIRS[3] | 55.43% | 94.78% |
| DIRS[1] | 87.8% | 99.26% |
| FWDIRS (Proposed System) | 93.23% | 98.66% |

As shown in Table 1, the average precision in WDIRS and DIRS is 55.43% and 87.8%, repectively. Also, average recall in WDIRS and DIRS is 94.78%. and 99.26%, respectively. After apply feature weighting method to DIRS the average precision is 93.23% and average recall become 98.66% respectively.

Furthermore, in order to test the robustness of the proposed method to the type of fonts, the query font was changed to ''Tahoma''.

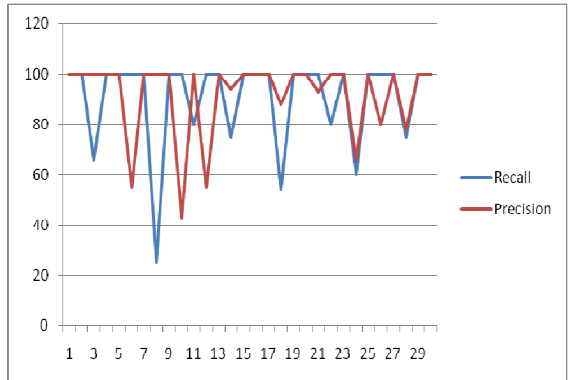

Fig 7. The variation of the precision and Recall Coefficient of the proposed FWDIRS for 30 searches with the query font "Tahoma". The Average Precision is 90.5% and the average Recall is 91.73%

As shown in Figure 7, by proposed method for feature weighting, performance of DIRS in term of average precision and Recall is increased.

Table 2 dpicts comparison the average precision and average recall of the approach with DIRS [1].

Table 2. Comparison the average precision and recall between proposed system and DIRS with the query font "Tahoma"

|  | Precision | Recall |
|---|---|---|
| DIRS[1] | 89.44% | 98.05% |
| FWDIRS (Proposed System) | 90.5% | 91.73% |

As shown in Table 1, the average precision in DIRS is 89.44% . Also, average recall in DIRS is 88.05%. After apply feature weighting method to DIRS the average precision is 90.5% and average recall become 91.73%.

6.1 Validating the Weighting Algorithm

To evaluate the proposed feature weighting method, we cluster storing data in database without using weighted features at first. In this approach, each word is a cluster. In this problem, the cardinality of cluster is unknown. Since, we propose Improved K-means algorithm (IK-Means) for clustering data. The Improved K-means algorithm uses a specified threshold for distance between words to create a new cluster. Improved k-means can be formally defined in the following steps:
1. Specify the cluster threshold to create new cluster.
2. create_new_cluster=true.
3. Assign d (1) to c (1).
4. For i form 2 to n
    a. For all clusters do
        i. Calculate distance (d (i), c (j)).
        ii. If(distance(d(i),c(j))≤threshold) assign d(i) to c(j) and create_new_cluster=false.
    b. If (create_new_cluster is true) then create a cluster, k=k+1, c (k) =d (i).
5. Run K-Means using the found clusters, and the full original dataset.

Where d (i) is i[th] record in database and c (j) is centroid of cluster j. after clustering the stored data in the database without using the weighted features, we clustering the data with using weighted features.

In order to test the performance of the weighted features based IK-Means algorithm, experiment results of both original IK-means algorithm and weighted features based IK-means algorithm, a data set with 10 data points was clustered in the experiment.

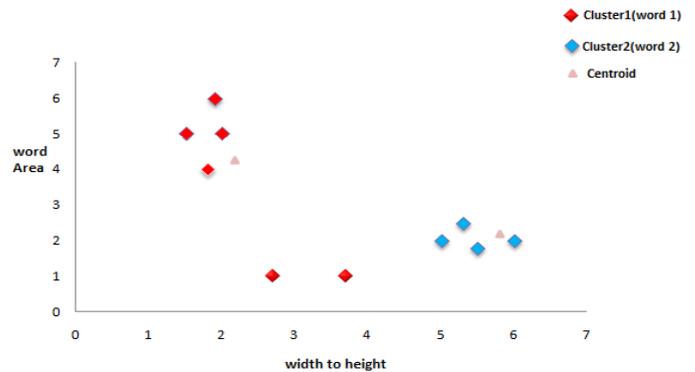

Fig 7. Clustering without using weighted features

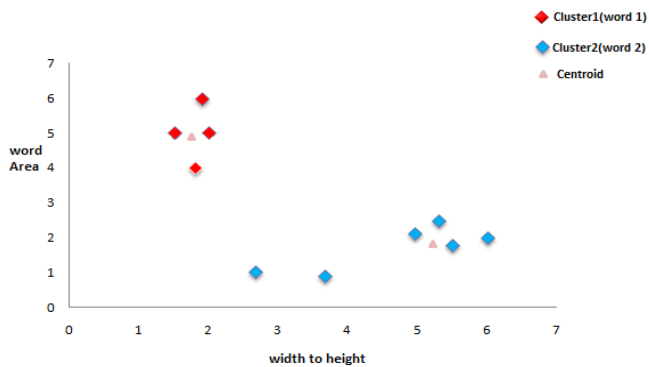

Fig 8. cluster with using weighted features

Figure 7 shows that 2 objects which are Cluster 2 are clustered to Cluster 1. Figure 8 shows that all objects are correctly clustered to Cluster 1 and Cluster 2 respectively. Moreover, we use Typical objective functions in clustering formalize the goal of attaining high intra-cluster similarity (words within a cluster are similar) and low inter cluster similarity (words from different clusters are dissimilar). This is an internal criterion for the quality of a clustering.

## 7. Conclusions

In this paper, we present feature weighting method in document image retrieval system. Proposed method use coefficient of multiple correlations for weighting each attributes. Purpose of this method is to increase performance of DIRS. The obtained results indicates that proposed method is able to particularly increase the performance of the DIRS in terms of average recall and average precision.

## References

[1] Konstantinos Zagoris, Ergina Kavallieratou, Nikos Papamarkos: "A Document Image Retrieval System". Eng. Appl. of AI 23(6), 2010,pp. 872-879.
[2] Y.C.Song, H.D.Meng, M.O'Grady, and G.O'Hare, 2007,"Applications of Attributes Weighting in Data Mining." In IEEE Proc. of SMC UK &RI 6th Conference on Cybernetic Systems, pp.41-45.
[3] K. Zagoris, N. Papamarkos and C. Chamzas," Web Document Image Retrieval System Based On Word spotting", ICIP,2006, pp.477-480.
[4] Million Meshesha · C. V. Jawahar, "Matching word images for content-based retrieval from printed document images", International Journal on Document Analysis and Recognition, Vol. 11, No. 1, 2008,pp.29-38.
[5] Shuyong Bai, Linlin Li and Chew Lim Tan," Keyword Spotting in Document Images through Word Shape Coding",10th International Conference on Document Analysis and Recognition,2009, pp.331-335.
[6] Shijian Lu, Linlin Li, chew lim tan, "Document Image Retrieval through Word Shape Coding", IEEE Transactions on Pattern Analysis and Machine Intelligence Volume 30 Issue 11, 2008, pp.1913-1918.
[7] David Doermann. "The Indexing and Retrieval of Document Images. A Survey". Computer Vision and Image Understanding (CVIU) 70, 1998,pp. 287-298.
[8] Leydier, Y., Le Bourgeois, F., Emptoz, H, Textual indexation of ancient documents DocEng'05, November 2–4, 2005, pp. 111–117.
[9] Yue Lu and Chew Lim Tan. "Information Retrieval in Document Image Databases". IEEE Tran. On Knowledge and Data Eng., vol.16, no. 11, 2004.
[10] Manesh B. Kokare and M.S.Shirdhonkar. Document Image Retrieval. An Overview. International Journal of Computer Applications (0975 – 8887) Volume 1 – No. 7, 2010, pp. 114-119.
[11] Marinai, S., Marino, E., Soda, G.,"Fontadaptative word indexing of modern printed documents". IEEE Trans. Pattern Anal. Mach.Intell. (PAMI) **28**(8), 2006, PP. 1187–1199.
[12] Tan, C.L., Huang, W., Yu, Z., Xu, Y. "Imaged document text retrieval without OCR". IEEE Trans. Pattern Anal. Mach. Intell.**24** (6), 2002, 838–844.
[13] Trenkle, J.M., Vogt, R.C. "Word recognition for information retrieval in the image domain". In: Symposium on Document Analysis and Information Retrieval,1993, pp. 105–122.
[14] Razvan Andonie, Angel Cataron, Lucian Mircea Sasu. **"**Fuzzy ARTMAP with Feature Weighting", Proceedings of the 26th IASTED International Conference on Artificial Intelligence and Applications, 2008, pp.91-96.
[15] D.S. Modha and W.S. Spangler, "Feature Weighting in k-Means Clustering". Machine Learning, Springer, 2003.
[16] K.Sparck Jones, "Indexing term weighting", Information Storage and Retrieval, vol. 9, 1973, pp. 619–633.
[17] J.Zhang and TN.Nguyen, "a new term significance weighting approach", Journal of Intelligent information system, vol. 24. No. 1, 2005, pp. 61-85.

**Mohammadreza Keyvanpour** is an Associate Professor at Alzahra University, Tehran, Iran. He received his B.s in software engineering From Iran University of Science &Technology, Tehran, Iran. He received his M.s and PhD in software engineering from Tarbiat Modares University, Tehran, Iran. His research interests Include image retrieval and data mining.

**Reza Tavoli** received his B.s in software engineering from Iran University of Science & Technology, Behshahr, Iran. He received his M.s software engineering from Islamic Azad University, science & Research Branch, Tehran, Iran. Currently, He is pursuing PhD in Software engineering at Islamic Azad University, Qazvin Branch, and Qazvin, Iran. His research interests Include document image retrieval and data mining.